\documentclass[journal]{IEEEtran}

\usepackage{graphicx}
\usepackage[export]{adjustbox}
\usepackage{float}

\usepackage{amsmath}
\usepackage[noadjust]{cite}
\graphicspath{ {./images/} }
\usepackage{float}
\usepackage{caption}

\begin{document}
%
\title{\textbf{Automatic Neuronal Activity Segmentation in Fast Four Dimensional Spatio-Temporal Fluorescence Imaging using Bayesian Approach}}
%
%
%

\author{\IEEEauthorblockN{Ran Li\IEEEauthorrefmark{1}\IEEEauthorrefmark{3},
Pan Xiao\IEEEauthorrefmark{1}\IEEEauthorrefmark{3}, Kaushik Dutta\IEEEauthorrefmark{1}\IEEEauthorrefmark{3} and Youdong Guo\IEEEauthorrefmark{2}\IEEEauthorrefmark{3}\thanks{Corresponding author: youdong.guo@wustl.edu}}\\
\IEEEauthorblockA{\IEEEauthorrefmark{1}
Department of Radiology\\ 
}
 \IEEEauthorblockA{\IEEEauthorrefmark{2}
Department of Neuroscience\\ 
Washington University in St Louis School of Medicine
 }

\IEEEauthorblockA{\IEEEauthorrefmark{3} \textbf{All authors have equal contributions}}
}

\maketitle

\begin{abstract}
Fluorescence Microcopy Calcium Imaging is a fundamental tool to \textit{in-vivo} record and analyze large scale neuronal activities simultaneously at a single cell resolution. Automatic and precise detection of behaviorally relevant neuron activity from the recordings is critical to study the mapping of brain activity in organisms. However a perpetual bottleneck to this problem is the manual segmentation which is time and labor intensive and lacks generalizability. To this end, we present a Bayesian Deep Learning Framework to detect neuronal activities in 4D spatio-temporal data obtained by light sheet microscopy. Our approach accounts for the use of temporal information by calculating pixel wise correlation maps and combines it with spatial information given by the mean summary image. The Bayesian framework not only produces probability segmentation maps but also models the uncertainty pertaining to active neuron detection. To evaluate the accuracy of our framework we implemented the test of reproducibility to assert the generalization of the network to detect neuron activity. The network achieved a mean Dice Score of 0.81 relative to the synthetic Ground Truth obtained by Otsu's method and a mean Dice Score of 0.79 between the first and second run for test of reproducibility. Our method successfully deployed can be used for rapid detection of active neuronal activities for behavioural studies.
\end{abstract}

\begin{IEEEkeywords}
Calcium imaging, Bayesian Deep Learning, Cell segmentation
\end{IEEEkeywords}

\IEEEpeerreviewmaketitle

\section{\textbf{Introduction}}

\IEEEPARstart{C}{alcium} Imaging has emerged as a powerful tool for understanding and analysing the function of large scale neuronal activities in the brain of biological specimens \cite{dombeck2007imaging}. These neuronal activities causes brief changes in the intra-cellular concentration of genetically encoded calcium binding indicators which are detected by advanced optical techniques \cite{baker2005imaging}. Traditionally laser scanning microscopy (LSM), such as two-photon and confocal microscopy was utilised to collect fluroescence signals \cite{denk1990two, helmchen2005deep}. Unfortunately, the LSM suffers from slow acquisition over fields with multiple neurons resulting in inadequate temporal resolution for recording large timescale of rapid neuronal activity. This leads to tradeoff between the signal-to-noise ratio(SNR), temporal resolution and scanned field of view which can be mitigated when multiple pixels are sampled in parallel. The light sheet microscopy which utilizes thin illumination sheet coupled to objective and aligned with the focal plane can achieve rapid optical sectioning exhibiting high SNR at 100 times higher pixel rates than LSM \cite{holekamp2008fast}. \\
\\
\begin{figure}
    \centering
    \includegraphics[scale=0.5]{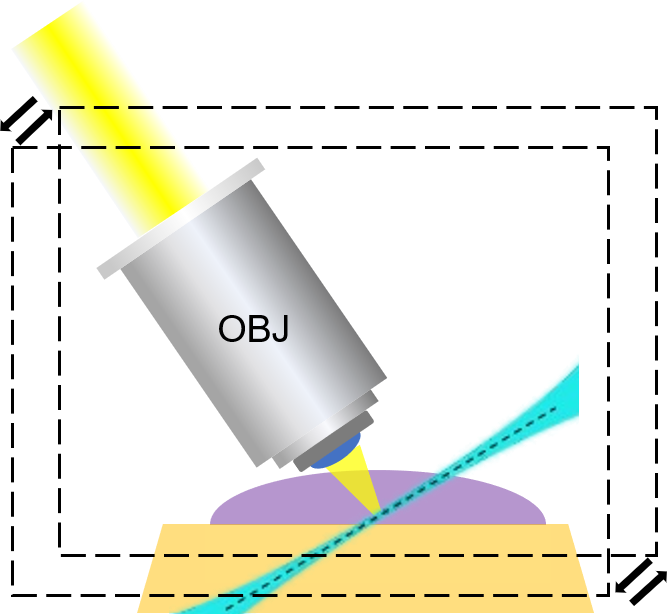}
    \caption{Objective-Coupled Planar Illumination (OCPI) microscopy}
    \label{fig:my_label}
\end{figure}
\begin{figure*}[h]
    \centering
    \includegraphics[width=\textwidth,height=10.5cm]{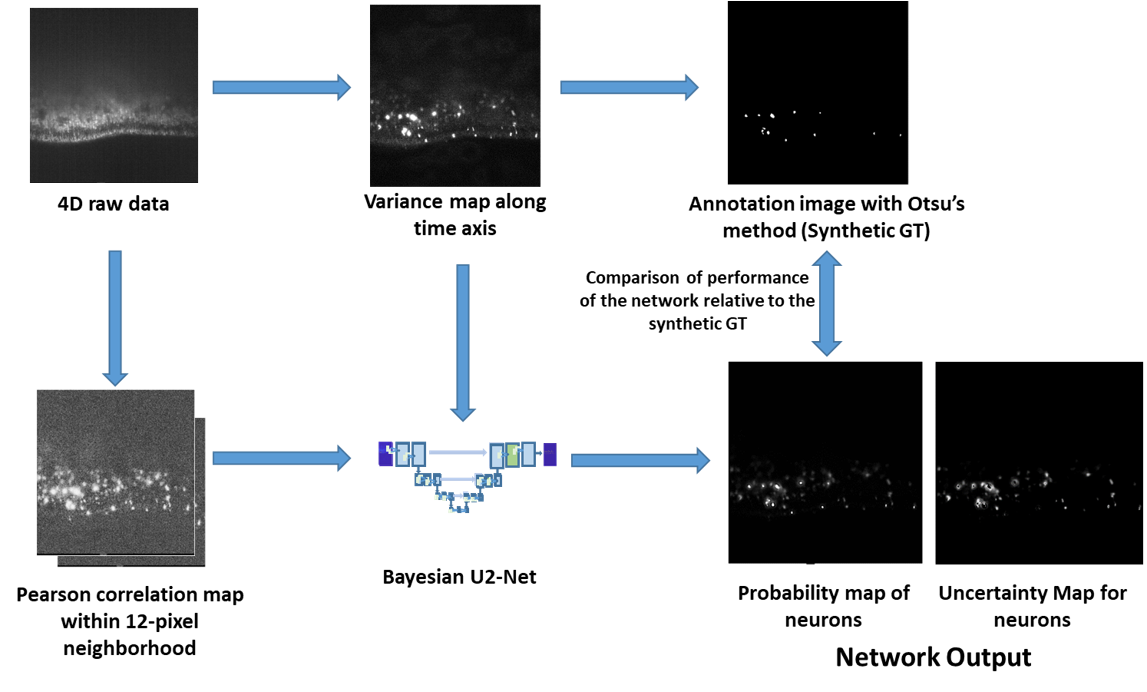}
    \caption{Overview of the Pipeline}
    \label{fig:my_label}
\end{figure*}
Conventionally, a human expert was employed to annotate regions of interests (ROI) corresponding to neuronal activities which was followed by summing each frame of the video and a subsequent deconvolution in temporal dimension. With large imaging datasets manual annotation becomes labor intensive and suffers from lack of reproducibility. Moreover, manual methods cannot de-convolve spatially overlapping signals. Thus an automatic pipeline is essential for decoding of neuronal activities in behavioral studies. The methods to detect of cells in calcium imaging extends from traditional unsupervised machine learning \cite{maruyama2014detecting,pnevmatikakis2013sparse,pnevmatikakis2014structured,diego2014sparse,pachitariu2017suite2p,reynolds2017able,giovannucci2019caiman,friedrich2017fast,inan2017robust,mukamel2009automated} and clustering \cite{spaen2019hnccorr,kaifosh2014sima} to modern deep learning approach \cite{apthorpe2016automatic,klibisz2017fast,xiao2025sc,soltanian2019fast,kirschbaum2020disco}. Earlier automatic cell segmentation methods employed classical machine learning techniques like principal component and independent component analysis (PCA/ICA) \cite{mukamel2009automated}, level set segmentation \cite{reynolds2017able}, non-negative matrix factorization (NMF) \cite{pachitariu2017suite2p, maruyama2014detecting,  guo2024gsvd} and constrained form of NMF \cite{pnevmatikakis2013sparse, pnevmatikakis2014structured, guo2025optimal}. Giovannucci et al. \cite{giovannucci2019caiman} developed a faster and more scalable version of CNMF by employing a convolutional neural network and new initialization methods called CalmAn and CalmAn Online. Diego et al. \cite{diego2013automated} used sparse approximation coding in sparse temporal and spatial space to detect relevant activity. Spaen et al. \cite{spaen2019hnccorr} developed HNCCorr algorithm which uses a distance based combinatoral optimization approach to define correlation between pixels and divide them by partitioning algorithm. HNCCorr is also combined with shape based 2D convolutional network (Conv2D) for better detection of weak and constant neuronal activity.\\
\\
Deep Learning has emerged as an established practice in medical image segmentation due to its high capability of feature extraction and pixel level classification. Klibisz et al. \cite{klibisz2017fast} used the U-Net architecture to segment neurons from 2D summary images(calculated by taking mean across all frames) along the temporal dimension. Another CNN based approach implemented a (2+1)D network where 2D kernel is applied to each frames followed by aggregation of temporal information in higher network level \cite{apthorpe2016automatic}. Due to temporal approximation the 2D CNN methods are inadequate in identifying  sparsely firing or distinguish overlapping neurons. The STNeuroNet \cite{soltanian2019fast} utilizes a 3D CNN for exploiting the spatio-temporal dynamics from calcium imaging videos. Most recently Kirschbaum et al. \cite{kirschbaum2020disco} developed the DISCo method which computed the correlation maps on multiple segments along the temporal dimension and aggregated that information to summary image (spatial information) followed by a 3D spatiotemporal convolution to create affinities. An edge weighted graph is used to extract the neurons by partitioning from predicted affinities.\\
\\

The objective of this study was to develop a 2D CNN network based on Bayesian Uncertainty to segment active neuronal activities from 4D spatiotemporal data obtained from light sheet microscopy. In order to minimise the labor intensive task of manual segmentation we used a consensus based thresholding approach to make decision regarding the ground truth for training and validation of the network. To account for the temporal information from the calcium imaging videos a pixel wise correlation was implemented. The spatial and temporal information were fused into a tensor and served as the input for the Bayesian network. We validated the performance of the network by evaluating the reproducibility of the network.

\section{Method}
\subsection{Data Acquisition and Dataset Description}
The data is provided by Holy lab \cite{lee2019sensory}, intact mouse vomeronasal sensory epithelium was imaged by the lab built light sheet Objective-Coupled Planar Illumination (OCPI) microscope system. $620$ $\mu$m $\times$ $640$ $\mu$m $\times$ $300$ $\mu$m imaging volumes were acquired at 0.5 Hz. Each stimulus was presented once for 5.5 seconds in each stimulus block. There were 4 stimulus blocks total, each presenting the ligands in different pseudo-randomized order interleaved with the other stimuli. In this dataset, we used mice expressing GCaMP5g (and PAmCherry). The calcium imaging data is obtained in the format of $Width$ $\times$ $Height$ $\times$ $Blocks$ $\times$ $Frames$ after motion correction. We obtained a total of $248$ blocks of recordings for neuronal activity expression of the auditory cortex in mice. Each block consists of $60$ frames with $512 \times 512$ pixels per frame.

\begin{figure*}[]
  \includegraphics[width=\textwidth,height=7.5cm]{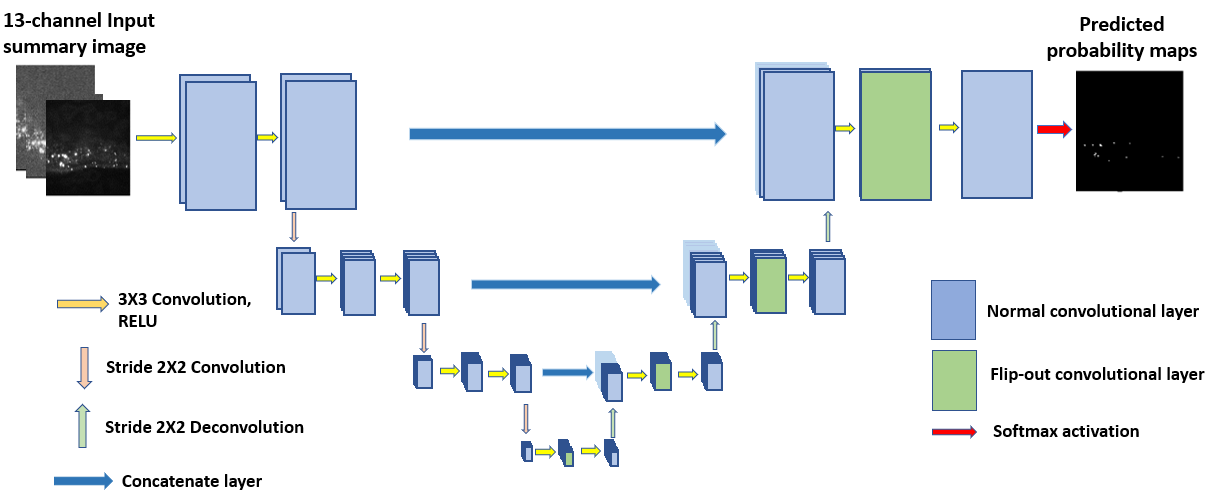}
  \caption{Bayesian Uncertainty U-Net Architecture}
  \captionsetup{justification=centering}
  \centering
  \label{fig:bayesian_s}
\end{figure*}

\subsection{Synthetic Ground Truth Generation by Adaptive Thresholding}
Deep Learning based segmentation requires a large amount of annotated ground truth for training the network. For vast majority of the calcium imaging data sets there is no solid ground truth which exists as delineating ground truth is labor and time intensive and suffers from inter-rater variability.  To generate large amount of synthetic ground truth required for training of our network, we used the method of Otsu's method to detect the neurons in the training datasets. The Otsu's method is used to perform automatic image thresholding \cite{otsu1979threshold} based segmentation where the algorithm return a single intensity threshold to separate pixels into binary classes. The threshold is determined by minimizing the weighted intra-class intensity variance. To make the neurons in the training data obvious and easy to segment, we generated the variance maps with respect to temporal frames of each individual neuron block and applied Otsu's method on those variance maps.

\subsection{Temporal information Extraction by Correlation}
Considering the huge differences between the fluorescence dynamics cells and background pixels, the temporal context from calcium imaging videos is crucial for detecting cells. We use the Pearson correlation coefficient \cite{pearson1895notes} to impose the temporal information from calcium imaging videos.

Suppose a video $\mathbf{X}\in \mathbf{R}^{T\times H\times W}$ with $T$ time frames and $H\times W$ pixels. Let the vector $\mathbf{x}_{hw}$ represent the signal of pixel $(h,w)$ of length $T$ with $\mathbf{x}_{hw}(t)=\mathbf{X}_{thw}$.  The Pearson correlation coefficient between the signals of two locations $(h,w)$ and $(h',w')$ is computed as:
\begin{align}
    c(\mathbf{x}_{hw},\mathbf{x}_{h'w'}) = \frac{\langle \mathbf{x}_{hw}-\bar{\mathbf{x}}_{hw}, \mathbf{x}_{h'w'}-\bar{\mathbf{x}}_{h'w'} \rangle}{\left \| \mathbf{x}_{hw}-\bar{\mathbf{x}}_{hw}\right \|_{2}\cdot \left \| \mathbf{x}_{h'w'}-\bar{\mathbf{x}}_{h'w'}\right \|_{2}}
\end{align}
where $\langle\cdot,\cdot \rangle$ denote the dot product, $\bar{\mathbf{x} }_{hw}$ and $\bar{\mathbf{x} }_{h'w'}$ represent the mean signal over time at the location $(h,w)$ and $(h',w')$ respectively.
The Pearson correlation coefficient can measure the linear correlation between the two time series. This measurement always has a value between $-1$ and $1$, $1$ for perfectly correlated signals and $-1$ for anti-correlated signals.

\subsection{Neural Network Architecture}
\subsubsection{Networks}
In consideration of the limitation of training data size and the accuracy of ground truth, we introduce a Bayesian approach to our neural network. Recent research on Bayesian neural networks demonstrates the feasibility of estimating uncertainty due to lack of data and perfect ground truth. The idea behind Bayesian Deep Learning is interlinked to the original Bayes theorem. The main objective is to study the posterior probabilities of the model parameter distributions of the network given an initial distribution of prior assumptions of these model parameters. The prior on these model parameters are updated based on repeated iterations through the training input data(likelihood). Once the entire dataset is parsed for an optimal number of iterations, the distribution over the model parameters i.e. the posterior, captures the updated belief about the optimal set of parameters to represent the data.\\

Bayesian Deep Learning facilitates to compute the uncertainties by modelling the posterior distribution $p(W|X,Y)$ over the weights $W$ of the network, for a training dataset of $\textbf{X} = \{X_k\}$ and their corresponding labels $\textbf{Y} = \{Y_k\}$ and $k = 1,2,....,N$ signifying the number of frames. The test dataset $\textbf{X}$ has 13 components i.e. 12 temporal correlation maps and a mean variance spatial map. Practically, calculating the exact posterior probability is computationally intractable hence Monte Carlo approximation is applied to find an approximation $q(W)$. The variational distribution $q(W_i)$ for the $i^{th}$ convolutional layer is modelled by using Monte Carlo sampling method, called Flipout estimator. The flipout estimator \cite{Wen2018} implements a stochastic forward pass via sampling from the convolution layer kernel and bias posteriors, which carries out Bayesian estimation to parameters.The mean of the resulting samples are used as an estimate of the segmentation and the variance is interpreted as uncertainty of the network.\\
\newline
We proposed a Bayesian Uncertainty U-Net (B-U2Net) \cite{nair2020exploring} that is based on a modified version of the standard U-Net \cite{ronneberger2015u}. The network architecture comprises of an encoder and a decoder, connected with skip connections. The encoder is composed of 5 convolutional blocks (with 64, 128, 256, 512 and 1024 output channels each). The first encoder block has two $3\times 3$ convolutional layers with $1\times 1$ padding. The other four blocks used two $3\times 3$ convolutional layers with $2\times2$ and $1\times 1$ padding respectively, each of them followed by batch normalization and leaky ReLUs. The decoder consisted of 4 deconvolutional blocks with 512, 256, 128 and 64 output channels. These blocks were modified to use $2\times2$ padding transposed convolution followed by a convolutional layer. A final $1\times 1$ convolutional layer was used at the end to retrieve the final probabilities for background/foreground. In order to enable epistemic uncertainty estimation, we changed convolutional layer with flipout convolutional layer in each deconvolutional block. The structure of B-U2Net is demonstrated in Fig. \ref{fig:bayesian_s}.\\ 
 \newline
To aggregate the temporal and spatial information from the calcium imaging videos as much as possible, the inputs of neural network comprise two parts with 13 channels: the summary variance images in one channel representing the spatial component and Pearson's correlation images in the other 12 channels i.e the temporal component. The output of this network comprises the probabilistic map of fired neurons and the uncertainty map of estimation. The network was trained with a mixture loss function, which is given by:
\begin{equation}
\begin{aligned}
    Loss = \frac{1}{3} (Loss_{ce}+Loss_{dice}+Loss_{KLD})
\end{aligned}
\end{equation}
where $Loss_{ce}$ is coress entropy loss, $Loss_{dice}$ is Dice loss and $Loss_{KLD}$ is K-L divergence loss. Those three loss functions are given individually by
\begin{equation}
    Loss_{ce} = \sum_{pixel }[(y_{true}\log(y_{pred})+(1-y_{true})\log(1-y_{pred})],
\end{equation}
\begin{equation}
    Loss_{dice}=1-\frac{2\sum_{pixel} y_{true}y_{pred}}{\sum_{pixel} y_{trye}^2+\sum_{pixel}y_{pred}^2},
\end{equation}
\begin{equation}
    Loss_{KLD} = \sum_{pixel }y_{true} \log (\frac{y_{true}}{y_{pred}}),
\end{equation}

\subsubsection{Training}
The input summary images and correlation images were normalized channel-wise with min-max normalization, which allows all data to range in 0 and 1. In order to find the best hyperparameter combination, we split our data into 80\% training set and 20\% testing set. 
We tested different parameter settings and applied the validation loss to determine the best hyperparameter setting. Subsequently, we used this set of hyperparameters to train the networks on the complete 4-D calcium images training set. The used hyperparameters can be found in table \ref{tab:1}. All the neural network is established on Tensorflow 2.0 platform and trained with Nvidia RTX 2080 ti graphic card. \\
\begin{table}[h!]
    \centering
  \begin{tabular}{cccc}
\hline
& epochs& batch size&learning rate    \\
\hline
Uncertainty UNet& 60& 2&0.0005\\
\hline
\end{tabular}
    \caption{Hyperparameters}
    \label{tab:1}
\end{table}

\begin{figure*}[]
  \includegraphics[width=\textwidth,height=7cm]{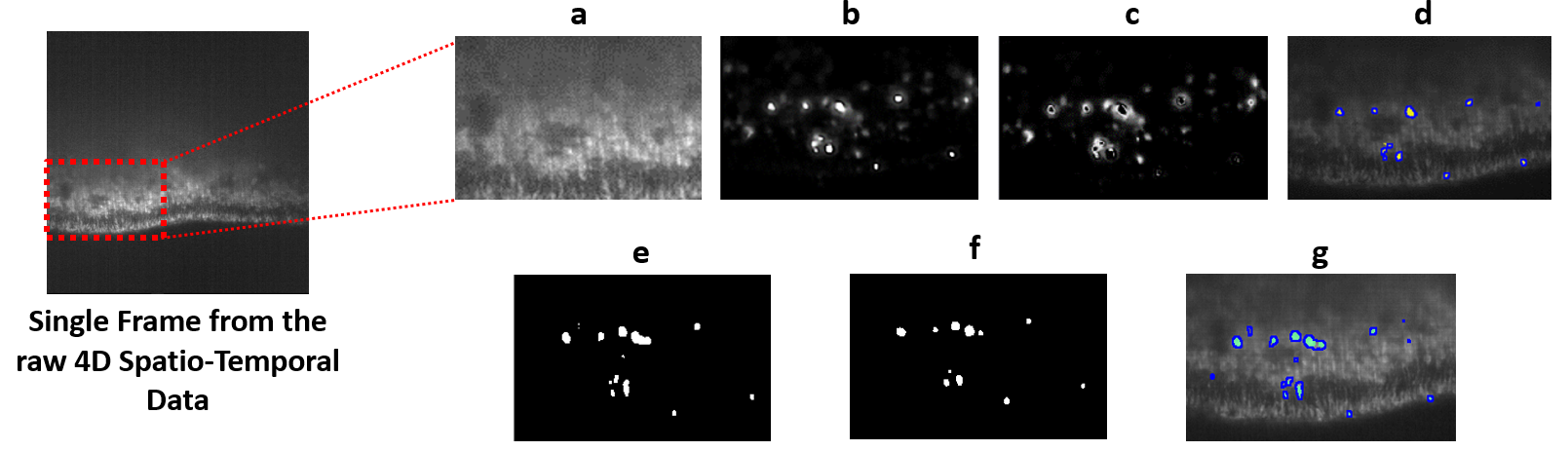}
  \caption{The depiction of the performance of the network represented on a single frame (a) Raw Image representing a frame from the 4D spatio-temporal data (b) Segmentation Probability Map obtained from the B-U2Net (c) Epistemic Uncertainty Map  of the network  (d) Overlaying of the results of the probability segmentation maps obtained by the network and the synthetic Ground Truth obtained by Otsu's Threshold on the representative frame  (e) Segmentation Map obtained when the network is trained with the first split of the dataset (f) Segmentation Map obtained when the network is trained with the second split of the dataset (g) The first and second segmnetation on the same representative frame is overlaid to signify the reproducibility of the algorithm}
  \captionsetup{justification=centering}
  \centering
  \label{fig:res}
\end{figure*}

\subsection{Evaluation Metrics}
To evaluate the performance of the probability maps generated by the Bayesian U2Net relative to the synthetically generated ground truth by Otsu's Methods we employed traditional metrics like Dice Score, Pixel Detection Accuracy and Sensitivity along with non-traditional parameters like Mathhews Correlation Coefficient (MCC) \cite{matthews1975comparison}. Though Dice Score and Accuracy are among the most popular metric for binary classification tasks, but they have the tendency to over optimistically inflate results specially for imbalanced data like we have in our case (i.e. the neuronal activity area is less than 10\% of the total region) \cite{chicco2020advantages}. As we had empirically thresholded synthetic ground truth and not expert segmented groundtruth so along with traditional segmentation evaluation we also tested the generalization and consistency of the network to produce reproducible results. To test the reproducibility the training dataset of 200 blocks were split into two parts. The B-U2Net was trained with individually with the two split training dataset under similar hyperparameters and tested on the same testing dataset of 48 blocks. The performance of the network in terms of reproducibility was expressed with metrics : Dice coefficient, pixels accuracy, sensitivity and  MCC.  
The mathematical equations for these four evaluation metrics are given in $(6)-(9)$. 
\begin{align}
    Dice  = \frac{2 TP}{FN+FP+2*TP}
\end{align}
\begin{align}
    Sensitivity = \frac{TP}{FN+TP}
\end{align}
\begin{align}
    Accuracy = \frac{TP+TN}{N}
\end{align}
\begin{align}
    MCC = \frac{TP/N - S*P}{\sqrt{P*S(1-S)(1-P)}}
\end{align}
where $N = FN+FP+TP+TN$, $S = (TP+FN)/N$, $P = (TP+FP)/N$.
$TP$, $TN$, $FP$ and $FN$ denote true positive, true negative, false positive and false negative respectively. 


\section{Results and Discussion}
We evaluated the performance of the B-U2Net using the independent testing dataset of 48 blocks of calcium imaging videos. The model's performance was assessed in terms of segmentation accuracy relative to the synthetic ground truth and the model's power to produce generalized and reproducible results.

\begin{figure}[]
\includegraphics[scale = 3.5]{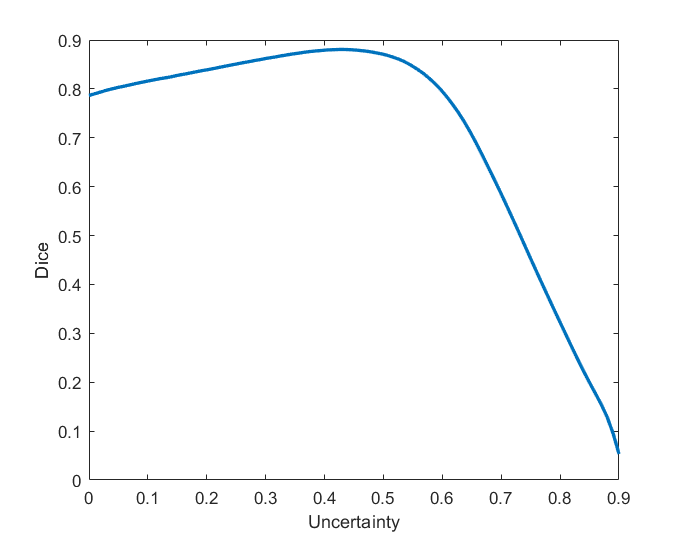}
\caption{Correlation between the Dice Score and Epistemic Uncertainty}
\captionsetup{justification=centering}
  \centering
  \label{fig:corr}
\end{figure}

\subsection{Semantic segmentation performance}
The parameters of Bayesian neural network are a bunch of posterior distributions conditioned on our training dataset. We ran an ensemble testing experiment including 40 testing predictors to evaluate our model completely, which can eliminate the randomness for one single predictor. The final probability map of neurons is the average of 40 outputs of predictors and the epistemic uncertainty map is the variance map with respect to the outputs of ensemble predictor. \\

In Table \ref{tab:2}, we showed the assessment outcome of segmentation performance with 95\% credible interval corresponding to different metrics. The equations of four performance evaluation metrics are shown in $(6)-(9)$. From Table \ref{tab:2} we observe that our model performs significantly well relative to the synthetic ground truth. The dice coefficient, sensitivity and MCC is around $0.8$. The accuracy is $0.97$ which means our model successfully detects $\sim80\%$ of the neurons and  successfully classifies  $>97\%$ of the background pixels.
\begin{table}[h!]
    \centering
  \begin{tabular}{ccccc}
\hline
& Dice& Accuracy&Sensitivity&MCC    \\
\hline
Semantic segmentation& 0.80& 0.97&0.79&0.81\\
\hline
\end{tabular}
    \caption{Semantic segmentation performance on 48 blocks testing data for Bayesian neural network.}
    \label{tab:2}
\end{table}

An representative result of the segmentation task performed by a model trained with our loss function is depicted for a single frame of 4D spatio-temporal data in Fig. 4. Fig. 4a signifies the portion of the calcium imaging frame which has clusters of neuronal activities. Fig. 4b represents the probability map of the active neurons in the portion as detected by the B-U2Net. Fig. 4c represents the epistemic uncertainty map associated to the probability segmentation. It's interesting to notice that some regions are dark with bright circle surrounding them. This phenomenon is resulted from that the segmentation results are in 95\% credible interval and the segmented regions in uncertainty map are dark with lower uncertainty. The bright regions in uncertainty map can also be taken as a reference to indicate the potential neurons. Fig. 4d shows the overlay image of ground truth and segmentation result, in which the blue circles are segmentation results and  the yellow dots are annotated neurons. We can observe that the contours of ground truth and segmentation results have significant overlapping to each other. Fig. 5 depicts the correlation between the B-U2Bet segmentation performance (as measured by using Dice coefficient) and the mean uncertainty for each of the blocks on the test set. The correlation illustrates that our method has a high Dice coefficient about 0.8 with 95\% credible interval.


\subsection{Evaluation of Reproducibility}
Reproducibility is an essential requirement for machine learning based methods.To show the generalizability and the independence on dataset of our model, a reproducibility experiment has been conducted. We separated our training dataset into two training set, and each training set has 100 blocks. Those two training sets were trained separately with our model  and the trained models were tested on the same testing 48 blocks. We evaluated the similarity of those two testing outputs with four metrics as mentioned in the previous part and the results are show in Table.\ref{tab:3}

\begin{table}[h!]
    \centering
  \begin{tabular}{ccccc}
\hline
& Dice& Accuracy&Sensitivity&MCC    \\
\hline
Semantic segmentation& 0.79& 0.96&0.77&0.8\\
\hline
\end{tabular}
    \caption{Performance of reproducibility testing.}
    \label{tab:3}
\end{table}

As we can see in Table. \ref{tab:3}, the similarity of two reproducibility testing output is 0.79 given by Dice coefficient, which indicates that our model has good generalization.  We can ensure the high independence of our model on dataset and parameters and its consistency and reliability has also been proven with this experiment.

\section{Conclusion}
In this paper we have presented an automated Bayesian learning based approach for detection of neuronal activities in four dimensional spatio-temporal data produced by light sheet microscopy. The core element our method is the formulation of active neuronal segmentation under Bayesian approximation. This approach along with producing a probability map also computes the uncertainty map which gives an measure of confidence on the detection of neuronal activity which are scattered in small clusters across the calcium imaging frames. Instead of relying on purely shape based summary images to be used as input to the network we provided the combination of segment-wise correlation maps and summary variance maps to feed the network for it to exploit both temporal and spatial features across the calcium imaging videos. Another unique approach of our work as compared to other methods is the elimination of the use of manual segmentation for generation of ground truth required for training of the network. We trained our network in empirical based Otsu's thresholding method and evaluated the generalization of the network by testing its reproducibility. Our framework was found to be significantly reproducible and thus can be deployed to detect clusters of large scale neuronal activities in real time behavioural studies.

\section*{Acknowledgment}
We extend our gratitude to Dr. Timothy Holy for providing us with the experimental 4D spatio-temporal data and also for his helpful discussions. We also thank  Dr.  Joseph  O’Sullivan  and  Dr.  Joseph  Culver for  providing us  constructive  suggestions  and  feedback throughout the execution of this project. We  also thank  the  Washington  University  Center  for  High  Performance Computing for providing computational resources for this  project. The  center  is  partially  funded  by  NIH   grants S10OD025200, 1S10RR022984-01A1 and 1S10OD018091-01.

\ifCLASSOPTIONcaptionsoff
  \newpage
\fi



%
\bibliography{references.bib}
\bibliographystyle{IEEEtran}

%








\end{document}